%% file: main.tex
\documentclass[10pt,twocolumn,letterpaper]{article}

\usepackage{iccv}
\usepackage{times}
\usepackage{epsfig}
\usepackage{graphicx}
\usepackage{amsmath}
\usepackage{amssymb}
\usepackage{gensymb}
\usepackage{graphicx}               
\usepackage{amsmath,amssymb}
\usepackage{verbatim} 

\makeatletter
\@namedef{ver@everyshi.sty}{}
\makeatother
\usepackage{pgf}
\usepackage{pgfplots}
\usepackage{tikz}
\usepackage{pgfplots}
\pgfplotsset{compat=1.14}
\usepackage{media9}
\usepackage{enumitem}
\usepackage{mathtools}  
\usepackage{xfrac} 
\usepackage{animate}
\usepackage{caption}
\usepackage[accsupp]{axessibility}  
\newcommand\tab[1][0.4cm]{\hspace*{#1}}

\usepackage[breaklinks=true,bookmarks=false]{hyperref}

\iccvfinalcopy 

\def\iccvPaperID{12} 

\ificcvfinal\fi

\begin{document}

\title{Precise Forecasting of Sky Images Using Spatial Warping}

\author{Leron Julian\\
Carnegie Mellon University\\
{\tt\small ljulian@andrew.cmu.edu}
\and
Aswin C.\ Sankaranarayanan\\
Carnegie Mellon University\\
{\tt\small saswin@andrew.cmu.edu}
}

\maketitle
\ificcvfinal\thispagestyle{empty}\fi

\input{abstract.tex}
\input{intro.tex}
\input{prior.tex}
\input{setup.tex}

\input{warp.tex}

\input{method.tex}
\input{results.tex}

\input{conclusion.tex}

\input{acknowledgments.tex}

{\small 
\bibliographystyle{ieee_fullname} 
\bibliography{biblo}
}

\end{document}

%% file: abstract.tex
\begin{abstract}
The intermittency of solar power, due to occlusion from cloud cover, is one of the key factors inhibiting its widespread use in both commercial and residential settings.
	%
 %
  Hence, real-time forecasting of solar irradiance for grid-connected photovoltaic systems is necessary to schedule and allocate resources across the grid.
   Ground-based imagers that capture wide field-of-view images of the sky are commonly used to monitor cloud movement around a particular site in an effort to forecast solar irradiance. 
However, these wide FOV imagers capture a distorted image of sky image, where regions near the horizon are heavily compressed.
This hinders the ability to precisely predict cloud motion near the horizon which especially affects prediction over longer time horizons.
%
  In this work, we combat the aforementioned constraint by introducing a deep learning method to predict a future sky image frame with higher resolution than previous methods.
  Our main contribution is to derive an optimal warping method to counter the adverse affects of clouds at the horizon, and learn a  framework for future sky image prediction which  better determines cloud evolution for longer time horizons.	
\end{abstract}

%% file: Intro.tex
\section{Introduction}	

\begin{figure}
\begin{center}
\centering
\footnotesize SkyNet-UNet\tab\tab PhyD-Net\tab\tab Optical Flow\tab\tab Truth

\includegraphics[width=0.475\textwidth]{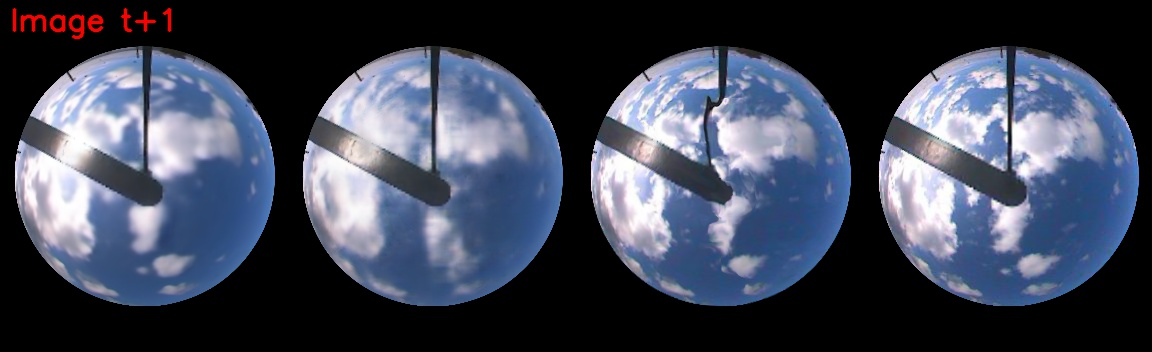}
\includegraphics[width=0.475\textwidth]{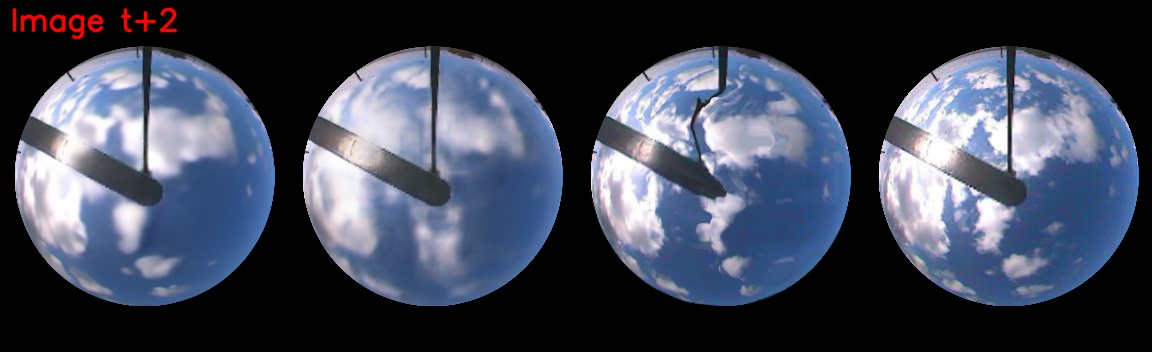}
\includegraphics[width=0.475\textwidth]{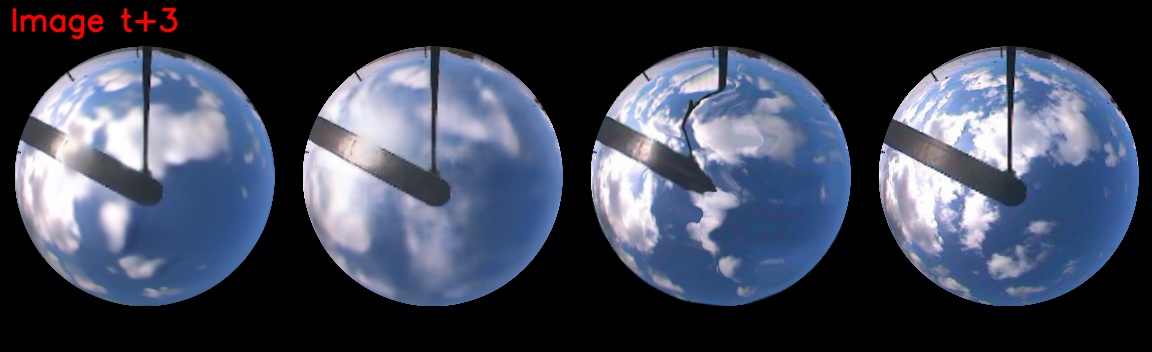}
\includegraphics[width=0.475\textwidth]{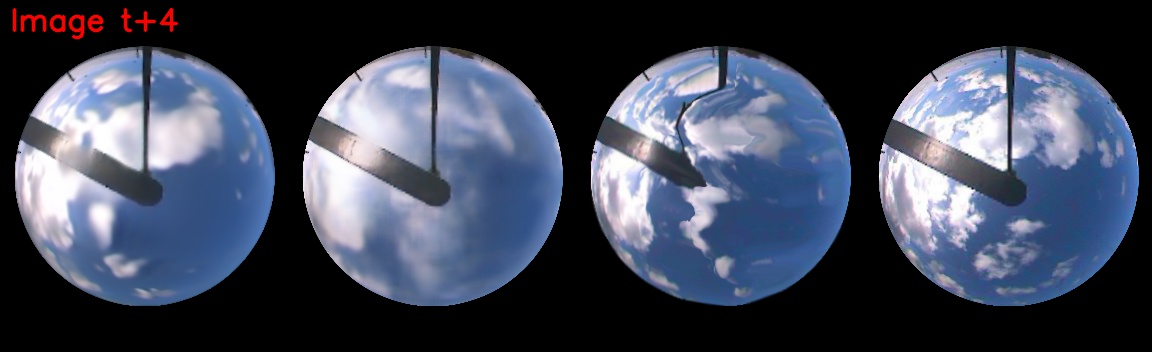}
\includegraphics[width=0.475\textwidth]{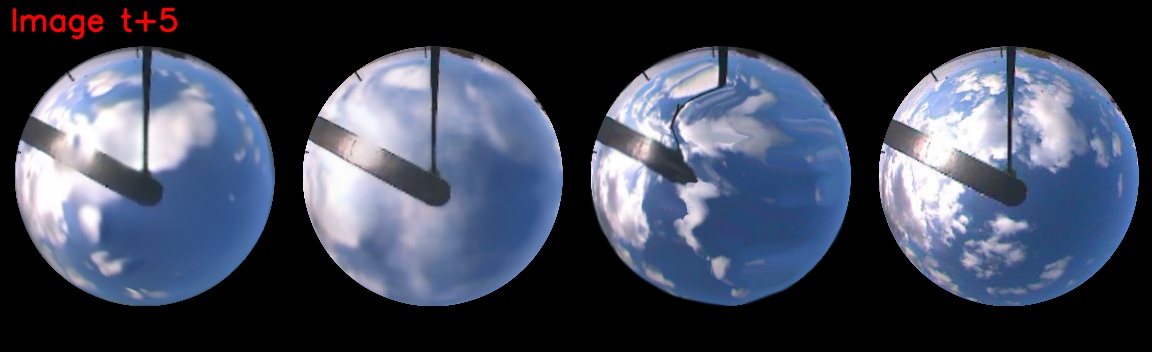}

\captionof{figure}{\textbf{Predicting future sky images.} Left-to-right: Results from SkyNet-UNet (the proposed technique), PhyD-Net-Dual \cite{9150809}, optical flow, and ground truth images. The methods take in as input images $[I_{t-5}, I_{t-3},I_{t-1},I_{t}]$, and predict future frames $[ \widehat{I}_{t+1}, \widehat{I}_{t+2},...,\widehat{I}_{t+5}]$; From top-to-bottom.}
\label{fig:Video_results}
\vspace{-2em}
\end{center}
\end{figure}

Solar irradiance, the output of light energy from the entire disk of the sun measured at a location on Earth, powers renewable photovoltaic energy systems for both residential and commercial power generation.
The amount of solar irradiance being received by these systems is highly influenced by cloud coverage that occludes, reflects, or scatters the rays directly.
Cloud cover, shape, thickness, and height are many variables that are difficult to predict and are unique day-to-day. 
As a result, solar energy is invariably intermittent, posing a significant challenge in its widespread usage \cite{Australia, SOVACOOL2009288}. This causes energy fluctuations and uncertainties that could lead to subsequent load balancing issues in power systems \cite{NBERw17086}.
Therefore, predicting solar irradiance for future time instances is essential for grid-connected photovoltaic systems to effectively schedule and allocate resources across the grid. In fact, by monitoring solar energy, power output can be optimized by utilizing a system of measured parameters to reconfigure the solar panel connection topology to improve efficiency and robustness in renewable energy systems \cite{Cyber-Physical}.	
	
One of the key approaches for predicting solar irradiance is by monitoring the movement of clouds around a particular site using  sky-images captured by the so called Total Sky Imager (TSI). 
In a typical TSI, a hemi-spherical image of the sky  with a 180$\degree$ field of view (FOV) is captured using a catadioptric system involving a camera observing the sky through a curved mirror; Figure \ref{fig:TSI_collage} provides examples of such images.
%
Most TSIs capture images periodically, say once every 30 seconds, and as a result, these images can be stacked together to create a time-lapse of historical cloud cover data around a particular site. 
Such TSIs,  coupled with a pyranometer that measures solar radiation as global horizontal irradiance (GHI), have been utilized in recent studies to nowcast and forecast solar irradiance \cite{7730887, electronics9101700, 2019arXiv190104881S}. 
Existing methods also use previous time instances of GHI to forecast future time instances of it using statistical models \cite{ALZAHRANI2017304, inbook}.  
More recent methods utilize deep neural networks to perform sky-image prediction for future time instances \cite{9258039} in which solar irradiance information, in the form of GHI, is extracted using various statistical or learning-based methods \cite{electronics9101700}.

Many of the initial works that model cloud movement in sky images utilize optical flow-based prediction; here, optical flow computed between current and past images is used to predict future sky images using simplistic modeling of cloud dynamics such as a constant velocity model.
Solely using optical flow for this application does not produce accurate long-term prediction due to the variability in velocity and amorphous shapes of clouds which makes forecasting their trajectory difficult. 
As a result, more recent works have focused on incorporating more sophisticated reasoning based on  deep neural networks to \textit{learn} to predict the future sky image.
%

The definitions of short and long-term prediction of cloud movement in sky-images is subjective to the sampling period, $T_0$, at which the images are captured. For this work $T_0$ is set as $T_0 = 30$ seconds. Therefore, short-term prediction is quantitatively defined as predicting 30 seconds in the future, whereas long-term prediction is anything greater than 30 seconds. Overall, however, both short and long-term prediction is difficult due to the constant changing shape of clouds.
This is exacerbated by distortions introduced by the hemispherical mirror used to capture the wide FOV sky-image; specifically, in a typical image obtained from a TSI, objects near the horizon are spatially-compressed and hence, appear much smaller at the horizon than when they are at the zenith. 
Due to this non-linear mapping produced by hemispherical mirrors, uniform physical motion of clouds leads to  apparent motion of varying magnitude on the image plane. 
Clouds at the zenith exhibit significantly larger apparent motion than clouds at the horizon. 
This in turn affects the accuracy of motion estimates for cloud movement tracking as the apparent motion at the horizon is extremely small and overhwelmed by the larger optical flow induced by clouds at the zenith.
For forecasting longer time horizons, the small movement of these clouds are what determines how the texture of clouds evolve over time. 
We attempt to counter this problem by warping the original image to a different space where the apparent motion is uniformly preserved both at the zenith and the horizon. 
This allows us to achieve longer forecasting times when modeling cloud evolution in sky images.

{{\flushleft \textbf{\emph{Contributions.}}}} In this work, we propose \textit{SkyNet}, which focuses on improving sky-image prediction. Our main contributions are as follows:
	 \begin{itemize}[leftmargin=*,noitemsep, topsep=-1pt]
	 		\item \emph{Cloud forecasting via spatially warped images.}  Our primary contribution is in showing that spatially-warping the sky images during training facilitates longer-forecasting of cloud evolution. This counters the adverse affects of resolution loss near the horizon.
	 	\item \emph{Incorporating larger temporal context.} We adapt prior work on future frame prediction in videos \cite{2017arXiv171209867L} to the case of sky images. Here, to increase precision in forecasting, we go beyond two input frames to usher in a larger temporal context. Specifically, to predict the image at time $t+1$, we take in as input four input images spanning $\{t-5, t-3, t-1, t\}$.
	 	\item \emph{Training and validation.} We train and evaluate our approach on a large dataset of sky images and demonstrate the ability to accurately forecast sky image frames with higher resolution metrics than previous cloud forecasting methods. We further use the forecasted sky-images to evaluate our results on estimating the GHI value for future time instances; Which is discussed in the Supplemental Material.
	 \end{itemize}	 
	 The accuracy of the proposed SkyNet predictions are shown in Figure \ref{fig:Video_results} where we show our ability to predict up to sky images for future time-instants to $t+5$. Each frame denotes a time lapse of 30 seconds in this dataset and hence, we predict up to two and a half minutes into the future.
	 

%% file: prior.tex
\section{Prior Work}
We discuss prior work in modeling cloud dynamics with the goal of  predicting solar irradiance.

\subsection{Modeling Sky Evolution Using Optical Flow}
Early works for cloud motion tracking such as those of  Ai \etal \cite{Ai2017AMO} and Jayadevan \etal  \cite{Jayadevan2012ForecastingSP} use a grid or block-based optical flow technique to model cloud velocity and motion. 
This optical flow method involves constructing a set of grid elements across the sky image in which the direction and velocity is then found between the correlation of grid blocks between adjacent frames. 
Although accurate for short-term cloud movement prediction, approximately 1 min, using this block based optical technique becomes increasingly difficult when complex cloud dynamics are involved. 
%
%
Therefore, this method becomes less accurate and tougher to forecast for long-term time horizons.

 Recent advances of optical flow techniques have improved upon block-based optical flow methods.
 Differential methods for optical flow estimation such as the Lucas-Kanade and Horn-Schunck  are common and popular techniques for estimating cloud motion \cite{8297111, 7455105, 8865296, 8662134}.

	
 Solely using optical flow to model cloud dynamics has immediate consequences due to the variability and constant changing of shape of clouds which makes forecasting their trajectory difficult. 
 More recent methods have seen better success by incorporating deep-learning methods, coupled with optical flow and other variables, to model cloud dynamics and evolution in the sky.
		
	\subsection{Modeling Sky Evolution Using Deep Learning}
Many of the works that utilize deep neural networks to model cloud dynamics predict a subsequent sky image for a future time instance using a convolutional neural network (CNN). 
This predicted sky image is then used to predict solar irradiance at that time instance.
	
Kato and Nakagawa \cite{9258039} use a convolutional long short-term memory network (LSTM)  with optical flow vectors and past sky images as input to generate a predicted sky image by extrapolating the flow vectors with the input images. 
Andrianakos \etal  \cite{DBLP:conf/iisa/AndrianakosTOKE19} utilize a generative adversarial network (GAN) for sky image prediction to counter the adverse blurry image effects of using traditional mean squared error loss (MSE) for image prediction. 
Le Guen and Thome \cite{9150809} incorporate physical knowledge in deep models based on PhyDNet \cite{2020arXiv200301460L} that exploits physical dynamics to enhance cloud motion modeling.	
	
Deep neural networks are currently the most recent methods for modeling cloud dynamics in sky image frames. 
However, precise forecasting of future sky image frames for longer time horizons is hindered by artifacts induced by the imager.
Fisheye camera lenses and hemispherical mirrors, commonly used for capturing sky images due to their wide angle FOV, compress the imagery near the horizon which affects the prediction of cloud evolution when forecasting sky images.
To counter this, in our work, we present a uniform warping scheme on the captured images to ensure that clouds further from the zenith of the hemispherical mirror have similar apparent motion to those in the periphery, so as to ensure accurate forecasting.

%% file: setup.tex
\section{Background and Problem Setup}
 We begin by describing how  sky images are captured along with the basic notation of how cloud occlusion relates to the amount of solar radiation being received at a site. 
 We follow this by deriving the proposed uniform warping method for  sky images.
	
\subsection{Total Sky Imagers and Solar Irradiance}
\begin{figure}[!ttt]
		\begin{center}
		 	\vspace{-40.5mm}
			 \includegraphics[width=0.475\textwidth]{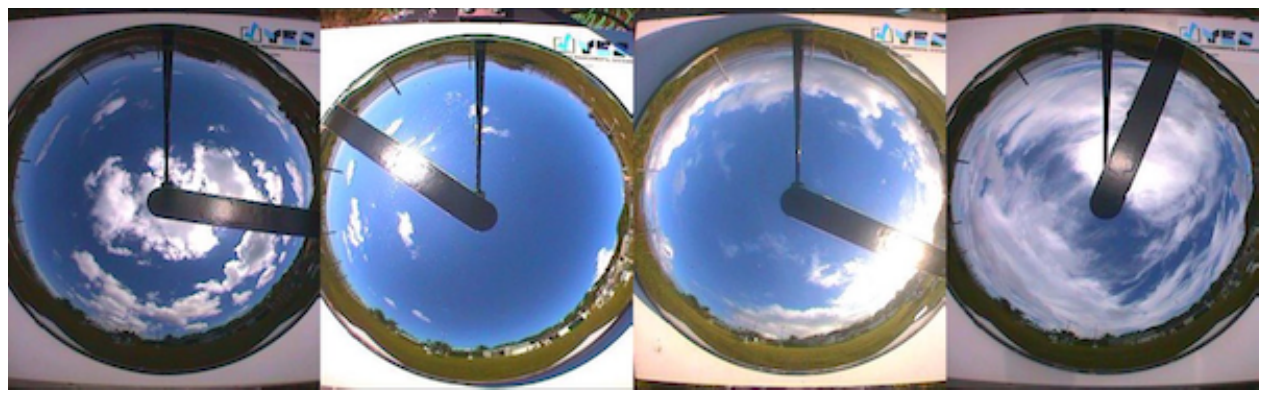}
			  \vspace{-48.5mm}
	              \caption{Sample images captured by a TSI \cite{TSI}.}
	              \label{fig:TSI_collage}
	            \end{center}
		\end{figure}

A TSI provides a time-lapse video sequence from an RGB camera that observes the sky via a hemispherical mirror \cite{TSI}. 
Generally, these systems are deployed to capture imagery of the sky at regular intervals for applications such as solar irradiance forecasting and visualizing cloud dynamics. 
To prevent damage of the camera sensor from direct expose of the sun, the TSI typically includes a mechanical arm that travels along the path of the sun throughout the images to occlude direct exposure. 
Figure \ref{fig:TSI_collage} shows some images from a TSI.

The RGB image captured from the TSI provides a sky map from which we can identify the location of clouds, their movement over time, and even a crude understanding of their absorption properties by associating the cloud cover at a time instance with it's associated GHI value when using a ground-based pyranometer.
%
%
Suppose that we have a solar panel collocated with the TSI, denoted by the location $x$. If the area of this panel is $A$ in $m^2$, then the radiant flux $\Phi(t)$ measured at time $t$ is given as:
\[ \Phi(t) = A \int_{\lambda} Q(\lambda) E_x(\lambda, t) d \lambda, \]
where $Q(\lambda)$ is the quantum efficiency of the panel, and $E_x(\lambda, t)$ is the spectral irradiance at the location of the panel, at the wavelength $\lambda$ and time $t$, expressed in the units of $J/(nm \cdot m^2)$.
This spectral irradiance can be related to the spectral radiance $L_x(\omega, \lambda, t)$ --- the flux at a point $x$ along a direction $\omega$ in the units of $J/(nm \cdot m^2 \cdot Sr)$. 
Therefore, the radiant flux $\Phi (t)$ can now be written as:
	\begin{equation}
		\resizebox{.9\hsize}{!} {$\Phi (t) = A \int_{\lambda \in \Lambda} Q(\lambda) \left[\int_{\omega \in \Omega}  L_x(\omega, \lambda, t) \max(0, \textbf{n}^{T} \omega) d \omega \right] d \lambda$}
	\end{equation}
	The set $\Omega$ defines the solid angle over which light is received at the solar panel and $\textbf{n}$ is the surface normal, or the orientation of the solar panel, in the same coordinates as $\omega$. 
	As an approximation, this integral (1) can be written as the occlusion map produced by the clouds multiplied by the spectral radiance due to sunlight as well as skylight which can be pre-measured.

\subsection{Problem Definition}
The goal of this paper is to provide a framework for short term prediction of the sky image.
Specifically, the TSI takes an image every $T_0$ seconds to provide a time lapse video.\footnote{For the dataset that we work with, this sampling period $T_0 = 30$ seconds. This choice balances the need to monitor fast moving clouds, that would benefit from shorter sampling period, and the size of the dataset, which scales inversely with $T_0$. }
For simplicity of notation, we denote this time-lapse video as a collection of frames $\{ \ldots, I_{t-1}, I_{t}, I_{t+1}, \ldots \}$, where $t$ is an integer-valued index for the sequence, keeping in mind that any two successive images are obtained $T_0$ seconds apart by the TSI.

Given $\{\ldots, I_{t-2}, I_{t-1}, I_{t}\}$, the past and current images in time lapse sequence at a time instant $t$,  our goal is to predict $\{ I_{t+1}, I_{t+2}, \ldots \}$, the images in the time lapse sequence for the next few instants.
Since clouds often move fast, there is little correlation between images taken at sufficiently far away time instances; hence, we can restrict the time horizon of images that we consider both for the input images (from the past)  as well as the predicted output images (of the future).
Hence, our objective can be refined to using the image set $\{I_{t-T_p}, \ldots, I_{t-1}, I_t \}$ to predict the image set $\{I_{t+1}, I_{t+2}, \ldots, I_{t+T_f}\}$, where the choice of the input time horizon $T_p$ and output time horizon $T_f$ are discussed later.

{{\flushleft \textbf{Challenges.}}} Modeling the evolution of the sky and predicting images at future time instants faces challenges that stem from the clouds themselves as well as features induced by the imager.
Clouds are amorphous, lacking the rich features that are prized in traditional motion modeling and flow estimation. 
Such domain-specific concerns can be handled by using learning techniques that implicitly build a prior for the underlying imagery.
However, even when using sophicated learning techniques, there are significant challenges that arise from the spatial distortions introduced by the TSI.

\begin{figure}[!ttt]
	\begin{center}
              \includegraphics[width=0.475\textwidth]{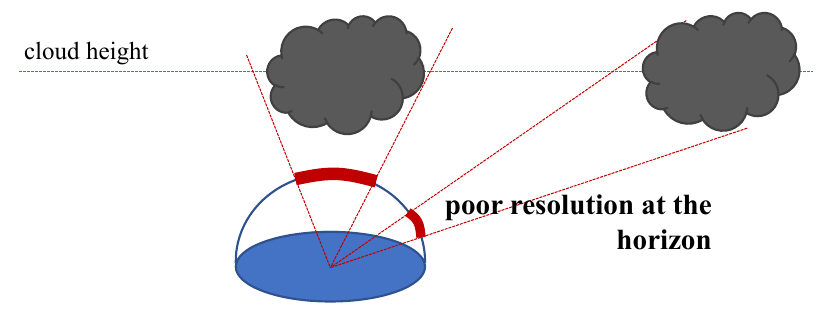}
              \caption{A cloud  subtends a smaller angle when it is further away from the zenith.  This results in the nonlinear spatial warping that is seen in Figure \ref{fig:TSI_collage}, and poses critical challenges for  effective forecasting of cloud movement. }
              \label{fig:CloudPosition}
            \end{center}
            \vspace{-1em}
	\end{figure}

Getting a $180^\circ$ FOV photograph with a TSI results in a highly nonlinear mapping between the sky and the image  as is seen in Figure \ref{fig:CloudPosition}. 
%
%
The effect of this distortion is easily seen in Figure \ref{fig:TSI_collage}. 
An immediate consequence of this nonlinear warping is that  motion near the horizon is not easily observable; for the same amount of cloud movement, the \textit{perceived optical flow} on the image plane of the camera is significantly smaller at the horizon.
This makes motion modeling near the horizon fragile to small perturbation.
This problem is  exacerbated by optical flow estimation, which is hard to perform on cloud imagery that lack high-contrast textures and the resulting flow estimates are inherently fragile, especially near the horizon.
While using enforcing smoothness priors on the flow estimates often leads to robustness especially at the zenith, they tend to make the flow at the horizon nearly zero.
Hence, the nonlinear spatial resolution is not conducive for predicting cloud evolution over longer time horizons.

{{\flushleft \textbf{Solution outline.}}} To address these challenges in motion estimation, and provide a framework for precise prediction of sky images, we make two modifications to traditional ideas in future frame prediction.
\begin{itemize}[leftmargin=*]
\item \textit{Optimal spatial warping.} First, under a simple model of image formation, we propose a warping of the TSI image so as to preserve motion of clouds over the spatial field. This serves to amplify motion near the horizon that is otherwise small. We describe this in Section \ref{sec:warp}.
\item \textit{Multi-image prediction.} Second, since the image after warping is  still smooth, we use multiple frames from the past to stabilize motion estimates. We perform this by adapting prior work on two-frame activity prediction. This is described in Section \ref{sec:method}.
\end{itemize}

%% file: warp.tex
\section{Optimal Warping of Sky Images}
\label{sec:warp}

 The warping of sky images is necessary because the apparent motion of clouds around the periphery of the hemispherical mirror will be much smaller than when at the zenith.
 %
 %
 As a result, we can only get good optical flow estimates at the zenith at the cost of poor optical flow estimates elsewhere. 
 For better long-term prediction of cloud evolution and as a result, better long-term prediction of solar irradiance, we spatially warp the images so that the apparent motion is more uniform. 
 We design a warping scheme so that over a specific site, we can achieve uniform optical flow.

\begin{figure}[!ttt]
              \includegraphics[width=0.475\textwidth]{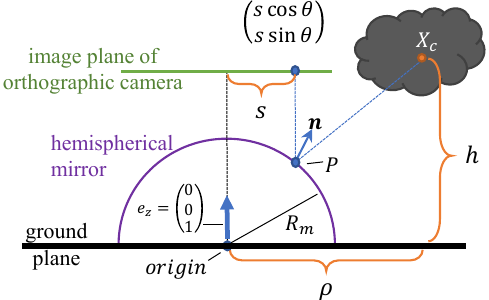}
              \caption{Overview of how the 3D position of a cloud in the world space gets mapped to a point on the image plane using a hemispherical mirror.}
              \label{fig:warp}
\end{figure}

{{\flushleft \textbf{Image formation model.}}} We model the imager as being an orthographic camera observing the sky through a spherical mirror of radius $R_m$.
The optical axis of the camera points is normal to the ground, and the optical center is aligned to the center of the spherical mirror.
We model the ground as being planar, an assumption that is reasonable given that the radius curvature of the earth is couple of orders of magnitude larger than the geographic region we can image with the TSI.
Given this, we adopt a world coordinate system whose origin is at the center of the spherical mirror.
The $xy$ coordinate plane is aligned with the ground plane and the $z$ axis is pointing towards the sky and hence, the optical axis of the camera is aligned to $[0,0,-1]^\top$.
Figure \ref{fig:warp} provides a schematic of this setup.

Suppose that a cloud at $X_c = [x_c, y_c, h]^\top$ maps to image pixel coordinates $[u_c, v_c]^\top$. 
We now seek to estimate the relationship between these quantities.
We first move from cartesian coordinates on the ground plane to polar coordinates, which allows us to exploit the rotational symmetry of the mirror about the $z$-axis.
With this, we can write the cartesian coordinates of the cloud as 
	\begin{equation}\label{eq:x_equation}
	X_c = [x_c, y_c, h]^\top = 
	\begin{bmatrix}\rho \cos \theta & \rho \sin \theta & h \end{bmatrix}^\top
	\end{equation}
and that of the image pixel coordinates as 
	\begin{equation}\label{eq:xpix_equation}
	[u_c, v_c]^\top = 
	\begin{bmatrix} s \cos \theta & s \sin \theta \end{bmatrix}^\top,
	\end{equation}
	where $(\rho, \theta)$ and $(s, \theta)$ are  polar cordinates for ground and image plane location, respectively, for the cloud.
	Note that we have effectively used the rotational symmetry of the mirror in insisting both the clouds and its corresponding camera pixel subtend the same angle $\theta$ in polar coordinates.
	

Let's denote $P$ as the point on the mirror that reflects the cloud to its corresponding image plane pixel.
Given that the camera is orthographic, we can derive the  $P$ to be
	\begin{equation}\label{eq:P}
	P = \begin{bmatrix} s \cos \theta & s \sin \theta & \sqrt{R_m^{2}-s^{2}} \end{bmatrix}^T
	\end{equation}
This comes from the fact that the point $P$ is on a sphere of radius $R_m$.
We can now enforce Snell's laws of reflection to relate $\rho$ and $s$ to each other, thereby getting the functional relationship between the  position of the cloud in world coordinates to its location on the image plane.
Specifically, we can write the surface normal at $P$, which is simply a unit norm vector oriented along $P$,  to be  equal to the average between the line produced by the point $P$ and the vertical line at $e_z$:
	\begin{equation}\label{eq:mirror_equation}
		\left( \dfrac{X_c-P}{\|X_c - P \|} + e_{z} \right ) \cdot \dfrac{1}{2} = \dfrac{P}{\| P \|}.
	\end{equation}
	Noting that $\| P \| = R_m,$ the radius of the sphere, we can express the 3rd coordinate of \eqref{eq:mirror_equation} as
	\begin{align*}
		\dfrac{1}{2} \cdot \left[ \dfrac{h - \sqrt{R_m^{2} - s^{2}}}{\| X_c - P \|} + 1 \right] 
	&=  \left[ \dfrac{\sqrt{R_m^{2} - s^{2}}}{R_m} \right].
	\end{align*}
	We can now solve for  $\| X_c - P \| $ to get
	$$
	 	\| X_c - P \|  = \gamma (s) = \frac{h - \sqrt{R_m^2 - s^2}}{\frac{2 \sqrt{R_m^2 - s^2}}{R} - 1} \approx \frac{h}{\frac{2 \sqrt{R_m^2 - s^2}}{R_m} - 1}
	 $$
	 
%
	 Now plugging $\gamma (s) = \| X_c - P \|$ back into \eqref{eq:mirror_equation}, we get
	 $$
	 	\dfrac{X_c- P}{\gamma (s)} + e_{z} = \dfrac{2P}{R_m},
	 $$
	 from which we can obtain an expression for $X_c$ as
	 \begin{equation}\label{eq:x_expanded}
	 	X_c = \gamma (s) \left[ \dfrac{2P}{\|P\|} + e_{z} \right] + P
	 \end{equation}
	 
	 Therefore, we can expand \eqref{eq:x_expanded}:
	 $$
	 	\resizebox{.9\hsize}{!} {$\begin{pmatrix} \rho \cos \theta \\ \rho \sin \theta \\ h \end{pmatrix} = \gamma (s) \left[ \dfrac{2}{R_m} \begin{pmatrix} s \cos \theta \\ s \sin \theta \\ \sqrt{R_m^{2}-s^{2}} \end{pmatrix} - \begin{pmatrix} 0 \\ 0 \\ 1 \end{pmatrix}  + \begin{pmatrix} s \cos \theta \\ s \sin \theta \\ \sqrt{R_m^{2}-s^{2}} \end{pmatrix} \right] $}
	 $$
	From the  top two rows of the previous equation, we can relate $\rho$ to $s$ as follows:
	\begin{align}
	 	\rho & =  \left(\dfrac{2 \gamma(s)}{R_m} +1 \right)s  =  \left(\dfrac{2 h}{2 \sqrt{R_m^2 - s^2} - R_m} +1 \right)s  \nonumber \\
	 	& \approx \frac{2 h s}{2 \sqrt{R_m^2 - s^2} - R_m}
	 \end{align}
 Instead of modeling $\rho$ directly, we can model $\frac{\rho}{h}$ which gives us height invariance 	 
	 \begin{equation}\label{eq:rho_tilde}
	 	\widetilde{\rho} = \frac{\rho}{h}  = \frac{2s}{2 \sqrt{R^2 - s^2} - R}
	 \end{equation}
	 Therefore, it does not matter the height at which the cloud is and therefore, we do not need to specify $h$.

{{\flushleft \textbf{Remarks.}}} While there is a significant distortion of the sky in the image plane of the camera, we can undo this distortion by redefining the image in terms of $\widetilde{\rho}$ instead of $s$.
That is, using the expression in \eqref{eq:rho_tilde}, we can map the image plane from $(s \cos\theta, s \sin\theta)$ to $(\widetilde{\rho}\cos\theta, \widetilde{\rho}\sin\theta)$.
This has the benefit of normalizing the observed optical flow so that it no longer suffers from the spatial distortion.
More specifically, in the transformed coordinates, the observed flow magnitude is the same, immaterial of where the motion occurs in the field of view of the device.

{{\flushleft \textbf{Inverting the warping function.}}} We can invert the relationship between $\widetilde{\rho}$ and $s$ in \eqref{eq:rho_tilde},  using some algebraic manipulation to get the following inverse relationship.
	 \begin{equation}\label{eq:s_quadratic}
	 	s = \frac{-R_m \widetilde{\rho} + R_m \tilde{\rho}  \sqrt{1 + 3(1 + \widetilde{\rho} ^2)}}{2(1+\widetilde{\rho}^2)}
	 \end{equation}

	\begin{figure}[!ttt]
	\begin{center}
              \includegraphics[width=0.475\textwidth ]{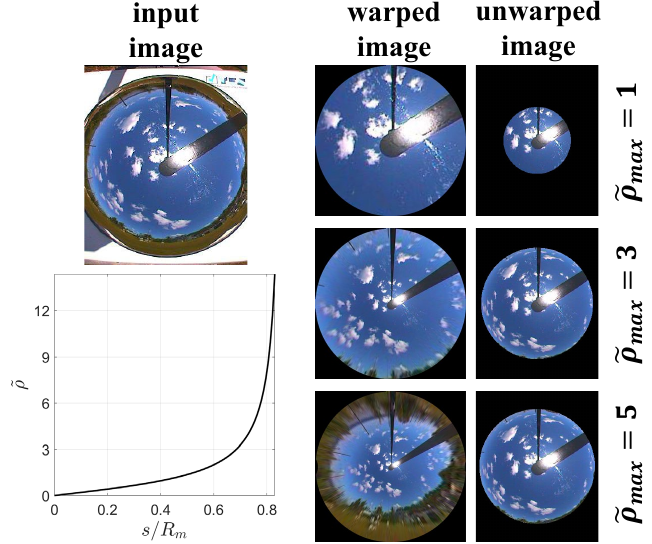}
              \caption{We show how the appearnce of an input image changes under the proposed warping. The plot on the left visualizes how we map from radial distances on the image to radial distances in the proposed representation. Choosing different values of the range of $\widetilde{\rho}$ produces different FOVs and associated distortions. This is visualized in the center column.  The right column shows the  image after inverting the warp to obtain the original image.}
              \label{fig:WarpSample}
            \end{center}
	\end{figure}
	
%
%
%
%
%
%
%

{{\flushleft \textbf{Implementation details.}}} To implement the warping function, we need to find the radius of the mirror $R_m$. We observe that for an orthographic camera and a planar ground, the horizon maps to a circle with a radius of $R_m/\sqrt{2}$. We use this to estimate $R_m$ in pixel count.
The other important parameter that we need to set is the maximum value of $\widetilde{\rho}$.
Setting $\widetilde{\rho} \in [0, \widetilde{\rho}_{\textrm{max}}]$ defines the FOV of the device to be restricted to $\pm \tan^{-1}\widetilde{\rho}_{\textrm{max}}$; for example, choosing $\widetilde{\rho} \in [0, 1]$ corresponds to a FoV of the sky of $\pm 45^{\degree}$.
 Figure \ref{fig:WarpSample} shows warped and unwarped images for different values of this range.
 A small value of this range leads to poor coverage of the sky and a large value has textures that are extremely blurred due to the nonlinearity of the warp as well as incorporation of trees and buildings.
For all of our experiments, we choose a range for $\widetilde{\rho} \in [0, 3]$ corresponding to a FoV of $\pm 71^\circ$ which provided a good balance between coverage and distortions.
Finally, to avoid loss of information, we upsample the image dimensions by a factor of three.

%
%
%
Figure \ref{fig:WarpSample} also shows how the warping can be inverted so as to  revert back to the original image space.
It is worth observing how the relative sizes of clouds at the zenith and horizon changes in the warped space.
This warping ensures that motion magnitude is preserved and spatially-invariant.
We next look at a learning framework for multi-image prediction.


%
%
%
%

%% file: method.tex
\section{SkyNet}
\label{sec:method}
To forecast sky images, we learn a deep neural network that we refer to as \textit{SkyNet} that takes in as input multiple images and predicts the next frame in the time lapse video.

	
	{{\flushleft \textbf{Network input.}}} As mentioned earlier, using multiple images to forecast provides us robust estimates of slow moving clouds as well as to combat the distortions introduced by the imager.
	To faciliate this, we use the stack of images $\{I_{t-5}, I_{t-3}, I_{t-1}, I_t\}$ to predict the image at $I_{t+1}$.
This choice reflects the need to have a long time horizon in the past, but given the redundancy, dropping some of the intermediate images help alleviate training time.
The images are warped using the approach described in Section \ref{sec:warp}.
	 
\begin{figure}
              \includegraphics[width=0.475\textwidth]{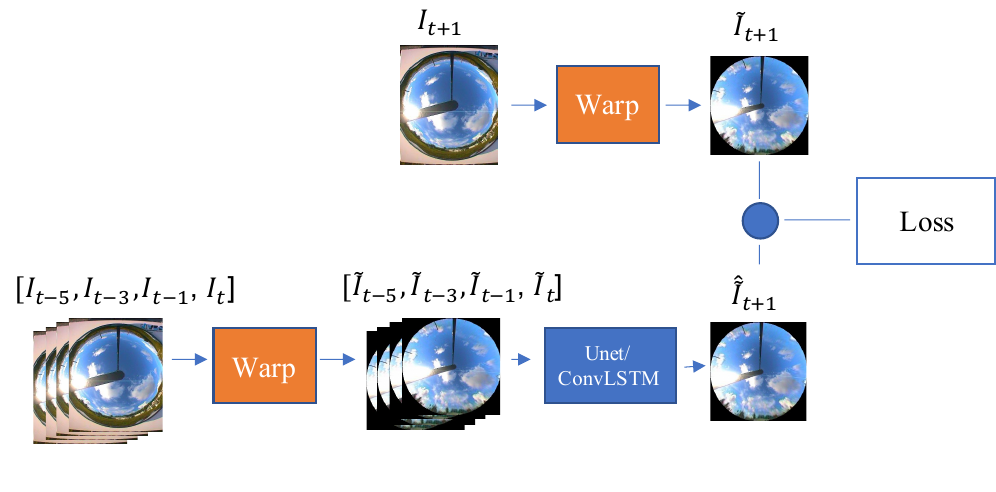}
              \caption{The proposed SkyNet forecasting method that incorporates the U-Net \cite{2015arXiv150504597R} or ConvLSTM \cite{2015arXiv150604214S} neural network architecture used to forecast a subsequent sky-image frame.}
              \label{Unet}
	\end{figure}

	{{\flushleft \textbf{Network architecture.}}}
	We consider 2 network architectures for our SkyNet Model.
	Initially, our first network architecture, SkyNet-UNet, adapts the future frame prediction model proposed for activity forecasting in  \cite{2017arXiv171209867L}. 
	The backbone of this architecture is a U-Net  \cite{2015arXiv150504597R} that takes in the input images to predict the image at the next time instant in the time lapse.
	Our second network architecture, SkyNet-LSTM, performs the same task as our initial network architecture, however, incorporates a convolutional long short-term memory network (ConvLSTM) \cite{2015arXiv150604214S}.
	Both architectures incorporate the same loss function further described below.
	Figure \ref{Unet} shows the structure of both forecasting models used.
	
	SkyNet-UNet starts with the number of input channels representing the number of time steps being considered. For each layer of the encoder, the number of channels are doubled until the bottleneck of the architecture which has 512 layers. The decoder, with skip connections between the encoder, brings the number of channels down to the target image size.
	
	SkyNet-LSTM incorporates a similar encoder-decoder architecture using 2 ConvLSTM cells for the encoder. 

{{\flushleft \textbf{Loss functions.}}}
The forecasting model enforces the predicted frames to be close to their ground truth in the spatial space as well as enforcing the optical flow between the predicted frames to be close to their optical flow ground truth as well. 
This is done by imposing a combination of penalties as network loss functions between the predicted frame $\widehat{I}_{t+1}$ and ground truth $I_{t+1}$. 
The network is trained using three loss functions based on intensity, gradient, and motion.
	The intensity loss ensures that pixels in the RGB space are similar by minimizing the $\ell_2$ distance between     $\widehat{I}$ and $I$:
	\begin{equation}
		L_{int}(\widehat{I}, I) = \| \widehat{I} - I \|^{2}_{2}
	\end{equation}
	When forecasting frames using the standard Mean Squared Error (MSE) loss function, the predicted images are blurry.
	 This is due to the fact that MSE generates the expected value of all the possibilities for each
pixel independently which causes a blurry Image. Therefore, the gradient loss is used to sharpen the predicted image:
	\begin{multline}
		L_{gd}(\widehat{I}, I) = \sum_{i, j} \| \mid \widehat{I}_{i, j} - \widehat{I}_{i-1,j} \mid - \mid I_{i, j} - I_{i-1,j} \mid \|_{1} \\ + \| \mid \widehat{I}_{i,  j} - \widehat{I}_{i, j-1} \mid - \mid I_{i, j} - I_{i, j-1} \mid \|_{1},
	\end{multline}
	where $i$ and $j$ are the spatial indices of the image.
	
	To predict an image with the correct motion, we place a loss on the optical flow field generated by the predicted image and the input image. 
In our work, a pre-trained CNN optical flow network is used \cite{2018arXiv180507036H} for the optical flow estimation. Denoting $f$ as the optical flow network used, the motion penalty is expressed as:
	\begin{equation}
		L_{op} = \| f(\widehat{I}_{t+1}, I_t) - f(I_{t+1}, I_t) \|_{1}
	\end{equation}

The three  functions  above are combined to define the  overall loss function as:
	\begin{multline}
		L = \lambda_{int} L_{int} (\widehat{I}_{t+1}, I_{t+1}) 
		+ \lambda_{gd} L_{gd} (\widehat{I}_{t+1}, I_{t+1}) + \\
	 \lambda_{op} L_{op}(\widehat{I}_{t+1}, I_{t+1}, I_t)
	\end{multline}
	We define $\lambda_{int} $, $\lambda_{gd} $, and $\lambda_{op} $ as $0.5$, $0.001$, $0.01$ respectfully.
	
{{\flushleft \textbf{Training Details.}}} Our implementation of the models are in Python using the PyTorch framework \cite{paszke2017automatic}. Training until convergence ends around 40 epochs with a learning rate of 0.001 using Adam optimization \cite{2014arXiv1412.6980K}. We run all of our experiments on 3 NVIDIA GeForce RTX 2080 Ti GPUs.
	
{{\flushleft \textbf{Long-Term Forecasting.}}}
	 To forecast a sky image frame longer into the future, we implement a simple recursive method. 
	 Once we have a prediction for $\widehat{I}_{t+1}$, to predict the image at time $t+2$, we use the image set
	 $\{ I_{t-4}, I_{t-2}, I_t, \widehat{I}_{t+1} \}$; that is, we use the predicted image at $t+1$ to recursively predict the next image in the sequence.
	 We can repeat this multiple times to increase the time horizon of the predictions.

%% file: results.tex
\section{Experiments}
\par We compare our method to prior deep-learning approaches to model cloud evolution in sky images along with the benefit of warping the sky images to achieve better long-term prediction.
		
\subsection{Sky-Image Dataset}
We use a publicly available dataset of TSI images for training and evaluation \cite{TSI}.
The source of the dataset is a TSI located on the Nauru Island and available for download at the Atmospheric Radiation Measurement facility.
Images in the dataset were captured over a duration spanning November 2002 to September 2013. Each successive image pairs are 30 seconds apart and are at a resolution of $352 \times 288$ pixels. 
In total, the dataset includes $4,272,938$ images. 
However, for our study, we utilize a subset of the available data as our primary train and test sets.
We utilize $42,171$ images from the year 2002 for training and validation and a disjoint set of $5,271$ images from 2003 for testing. Figure \ref{fig:TSI_collage} shows sample images from the dataset.

\begin{figure}
\begin{tikzpicture}
\pgfplotsset{
every axis legend/.append style={ at={(1.10,1.10)}, anchor=south east, legend columns = 2}}
\begin{axis}[width=0.475\textwidth,
    xlabel={Time Instance},
    ylabel={PSNR (dB)},
    xmin=0, xmax=6,
    ymin=27, ymax=37,
    xtick={1, 2, 3, 4, 5},
    xticklabels={$t+1$, $t+2$, $t+3$, $t+4$, $t+5$},
    ytick={27,29,31, 33, 35, 37},
    ymajorgrids=true,
    xmajorgrids=true,
    grid style=dashed,
]

\addplot[
    color=red,
    mark=o,
        line width=0.6mm,
    ]
    coordinates {
    (1,36.38)(2,34.49)(3,33.21)(4,32.43)(5,31.93)
    };

\addplot[
    color=blue,
    mark=square,
        line width=0.6mm,
    ]
    coordinates {
    (1,36.02)(2,30.94)(3,30.37)(4,30.37)(5,30.36)
    };

   \addplot[
    color=olive,
    mark=|,
        line width=0.6mm,
    ]
    coordinates {
    (1,35.29919721)(2,33.9817436)(3,33.21646999)(4,32.71272616)(5,32.33869024)
    };
    
     \addplot[
   dashed,
    color=green,
    mark=oplus* ,
        line width=0.6mm,
    ]
    coordinates {
    (1,32.01543404)(2,30.97217795)(3,30.21084355)(4,29.68936583)(5,29.32918954)
    };
    
    \addplot[
    dotted,
    color=black,
    mark=*,
    line width=0.6mm,
    ]
    coordinates {
    (1,31.97)(2,30.86)(3,30.18)(4,29.75)(5,29.44)
    };

    \addplot[
    dashed,
    color=purple,
    mark=l,
        line width=0.6mm,
    ]
    coordinates {
    (1,34.68)(2,33.13)(3,32.32)(4,31.9)(5,31.65)
    };

\addplot[
    color=orange,
    mark=x,
        line width=0.6mm,
    ]
    coordinates {
    (1,29.36)(2,28.88)(3,28.63)(4,28.49)(5,28.4)
    };

    \legend{SkyNet-UNet, SkyNet w/o warp, SkyNet-LSTM, SkyNet-LSTM w/o warp, PhyD-Net-Dual, 2-Input SkyNet, 2-Input SkyNet w/o warp}
 
\end{axis}
\end{tikzpicture}
\caption{Performance in image forecasting for various methods for a time horizon of one to five images.}
\label{fig:PSNR_Plot}

\end{figure}

%
%
%

\subsection{Comparison to Previous Methods}
Figure \ref{fig:Video_results} provides qualitative comparison between the SkyNet predictions, as well as basic optical flow-based prediction using a constant velocity model, and the PhyD-Net-Dual approach \cite{9150809}. 
As is seen in Figure \ref{fig:Video_results}, SkyNet predictions are of a significantly higher quality than the competitors.
We provide quantitative evaluation in the form of PSNR for these competing methods in Figure \ref{fig:PSNR_Plot}.
Here we also compare with a version of the SkyNet models without the optimal warping applied to it to study the influence that the warping function has.
We observe that there is a significant drop in performance when forecasting without the warping functions especially looking beyond the first forecasted image.

We also compare against a two-frame version of SkyNet-UNet, both with and without spatial warping, to test the effectiveness of using a larger time horizon.
In this version, we only provide $\{I_{t-1}, I_{t}\}$ as inputs to the network. 
As we expect, the performance of the prediction drops by a small amount when given a smaller past horizon, and by a larger amount when we disable spatial warping.

It should also be noted from Figure \ref{fig:PSNR_Plot} that although the SkyNet-UNet model performs the best at time instance t+1, as the images are forecasted longer into the future, the SkyNet-LSTM model outperforms all other tested models. This may be attributed to the long-term memory units of the convLSTM network however, further research will be done to solidify this argument.


\subsection{Dataset Size Dependent Results}

%
%
%
%
%
%
%
%

\par On a different training datasets and a new test dataset, we experiment with varying the size of the training data and asses how it influences our results. As shown in Table \ref{dataset_table}, as we increase the amount of training data from $1000$, $10000$, and $100000$ respectfully, the model performs better. This is due to the fact that more data allows the model to generalize better. At the same time, although we train on a small subset of the sky-image dataset, there are upwards of millions of images in total that can be utilized for training. Therefore, with the right amount of training, our method can be improved even further.

\begin{table}[ttt]
\begin{center}
\begin{tabular*}{0.475\textwidth}{cccccc}
\hline
 \multicolumn{6}{c}{PSNR values in dB} \\
 \hline
\hline
 & $\widehat{I}_{t+1}$ & $\widehat{I}_{t+2}$ & $\widehat{I}_{t+3}$ & $\widehat{I}_{t+4}$ & $\widehat{I}_{t+5}$ \\
\hline\hline
1K Dataset &31.24&31.05&30.94&30.88&30.84 \\
\hline
10K Dataset & 32.16	&31.71&31.4&31.19&31.05 \\
\hline
100K Dataset & 33.2	&32.64&32.28&32.03&31.84 \\
\hline

\end{tabular*}
\end{center}

\caption{Comparison of the peak signal-to-noise-ratio for various dataset size dependent results for 1K, 10K, and 100K sizes.}
\label{dataset_table}
\end{table}

%% file: conclusion.tex
\section{Conclusion and Discussions}
	\par In this work, we presented SkyNet which improved sky-image prediction to model cloud dynamics with higher spatial and temporal resolution than previous works. Our method handles distorted clouds near the horizon of the hemispherical mirror by patially-warping the sky images during training to facilitate longer forecasting of cloud evolution. Although our method performs well, the textures are still blurred near the horizon which is hard to undo and further degrades when predicting longer into the future. In future works, we plan to move away from the RGB image space and capture the 3D distribution of clouds. This will allow us to obtain a sense of the absorption and reflectance properties of clouds across a large scale to better attenuate how they affect the amount of solar radiation being received at the ground. Also, we would like to develop computational imaging approaches that captures wide-angle FOV images without the expense of objects being distorted near the horizon.
Above all, we believe our work is the first step toward precise prediction of solar irradiance to enable the widespread use of solar power both commercially and residentially.

{{\flushleft \textbf{\emph{Limitations.}} } Although SkyNet improves upon previous works modeling cloud dynamics, our method has limitations. 
	 First, due to the fact that we are using a learning-based algorithm, we are restricted to modeling clouds in the image intensity space where physical factors are not measured. 
	 Second, our model is also dataset dependent, inferring that sky images captured using a different camera than a TSI will require retraining on that camera specific dataset.

%% file: acknowledgments.tex
\section*{Acknowledgments}
This work was supported by the National Science Foundation under the CAREER award CCF-1652569. Leron Julian was also supported in part by the GEM fellowship and the Fritz Family Fellowship.

\newpage